\documentclass[sigconf]{acmart}

\usepackage{bbding}
\usepackage{algpseudocode} 
\usepackage{amsmath} 
\usepackage{graphicx}
\usepackage{subcaption}

\usepackage{multirow}
\usepackage{diagbox}
\usepackage{booktabs}
\usepackage{graphicx}
\usepackage{makecell}
\usepackage[ruled,vlined,linesnumbered]{algorithm2e}
\usepackage{bbding} 
\usepackage{pifont}
\usepackage{enumitem}
\usepackage{bbding}   

\SetKwProg{Fn}{Function}{:}{end}
\AtBeginDocument{%
  \providecommand\BibTeX{{%
    \normalfont B\kern-0.5em{\scshape i\kern-0.25em b}\kern-0.8em\TeX}}}

\copyrightyear{2026}
\acmYear{2026}
\setcopyright{cc}
\setcctype{by}
\acmConference[ICS '26]{2026 International Conference on Supercomputing}{July 06--09, 2026}{Belfast, United Kingdom}
\acmBooktitle{2026 International Conference on Supercomputing (ICS '26), July 06--09, 2026, Belfast, United Kingdom}
\acmDOI{10.1145/3797905.3807834}
\acmISBN{979-8-4007-2522-7/2026/07}

\ccsdesc[500]{Computing methodologies~Machine learning}
\begin{document}
\newcommand\blfootnote[1]{%
\begingroup
\renewcommand\thefootnote{}\footnote{#1}%
\addtocounter{footnote}{-1}%
\endgroup
}
\title{LayerScope: Predictive Cross-Layer Scheduling for Efficient Multi-Batch MoE Inference on Legacy Servers}


  

\author{Enda Yu}
\affiliation{%
  \institution{National University of Defense Technology}
  \city{Changsha}
  \country{China}
}
\email{yuenda@nudt.edu.cn}

\author{Dezun Dong}
\affiliation{%
  \institution{National University of Defense Technology}
  \city{Changsha}
  \country{China}
}
\email{dong@nudt.edu.cn}

\author{Zhaoning Zhang}
\affiliation{%
  \institution{National University of Defense Technology}
  \city{Changsha}
  \country{China}
}
\email{zhangzhaoning@nudt.edu.cn}

\author{Zhe Bai}
\affiliation{%
  \institution{National University of Defense Technology}
  \city{Changsha}
  \country{China}
}
\email{baizhe03o@nudt.edu.cn}

\author{Weiling Yang}
\affiliation{%
  \institution{National University of Defense Technology}
  \city{Changsha}
  \country{China}
}
\email{w.yang@nudt.edu.cn}

\author{Haojie Wang}
\affiliation{%
  \institution{Tsinghua University}
  \city{Beijing}
  \country{China}
}
\email{wanghaojie@tsinghua.edu.cn}

\author{Dongsheng Li}
\affiliation{%
  \institution{National University of Defense Technology}
  \city{Changsha}
  \country{China}
}
\email{dsli@nudt.edu.cn}

\author{Yongwei Wu}
\affiliation{%
  \institution{Tsinghua University}
  \city{Beijing}
  \country{China}
}
\email{wuyw@tsinghua.edu.cn}

\author{Xiangke Liao}
\affiliation{%
  \institution{National University of Defense Technology}
  \city{Changsha}
  \country{China}
}
\email{xkliao@nudt.edu.cn}

\begin{abstract}

Mixture-of-Experts (MoE) models reduce inference computation via sparse activation, but face a dual bottleneck—memory constraints and PCIe latency—when deployed on commodity or legacy hardware. For instance, deploying a 10B-parameter MoE on a single GPU requires offloading experts to CPU memory, where the resulting PCIe transfer latency often exceeds GPU computation by several times.
Hybrid CPU–GPU co-execution is promising, but confronts three intertwined challenges: inaccurate expert activation prediction, PCIe bandwidth contention between prefetch and on-demand loads, and high complexity in cross-device scheduling.
We present LayerScope, a prediction-driven dynamic expert scheduling system that addresses these challenges via three co-designed components. 
First, we find that expert activations exhibit distinct group-wise patterns across layers: near-input, near-output, and middle layers have unique characteristics. To address this, we propose the Learnable Layer-Aware Predictor (LLaPor), a module explicitly designed to achieve high-precision expert selection prediction by distinguishing and leveraging these layer-group-specific traits.
Second, we propose the Prefetch-Aware Cross-Layer Scheduling strategy (PreSched), a strategy that quantifies global performance gains, across all MoE layers, rather than confining to local optimality within individual layers, while intelligently balancing the trade-off between prefetching costs and on-demand loading overhead, and generates globally optimal scheduling plans.
Third, we design an Asynchronous I/O Optimizer (AsyncIO) that decouples I/O operations from computation, overlaps PCIe transfers with GPU/CPU kernel execution, and thus effectively eliminates idle waiting bubbles.
LayerScope achieves 141\% higher end-to-end inference throughput and 74.6\% lower decoding latency relative to state-of-the-art solutions. This breakthrough enables efficient deployment of large MoE models on commodity GPU-CPU platforms and legacy hardware, significantly reducing the cost barrier for MoE adoption by repurposing existing infrastructure in resource-constrained environments.

\end{abstract}

\keywords{Mixture-of-Experts; Offloading; LLM Inference}

\maketitle

\section{Introduction}

Mixture of Experts (MoE) \cite{moe,deepseekv2,moonlight,Qwen3} models boost computational efficiency for large language models via sparse activation, yet they encounter a critical memory bottleneck when deployed in resource-constrained  low-end commodity hardware and legacy devices widely retained in production environments \cite{BigMac,deepspeed,pagedattention}. 
For instance, loading all experts of Mixtral-8x7B \cite{moe} demands 80GB of memory, which far surpasses the 24GB capacity of mainstream commodity GPUs like the NVIDIA RTX 4090 and is even more prohibitive for aging enterprise-grade GPUs (e.g., NVIDIA Tesla V100 or older generations) with limited VRAM and outdated I/O interfaces, highlighting the urgency of MoE deployment solutions that support hardware reuse.

\begin{figure}[t]
  \centering
    \includegraphics[width=\linewidth]{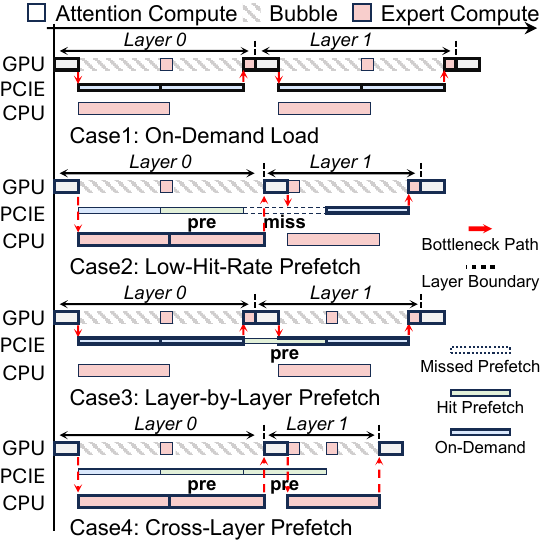}
    \caption{Impact of prefetching on inference latency.}
    \label{figure1} 
\end{figure}


Expert offloading \cite{hybridmoe,daop,heterollm} has become the primary approach to alleviate this memory bottleneck: experts are stored in CPU memory or SSDs \cite{klotski,InstAttention,Zero-infinity} and loaded into GPU on demand. This inevitably introduces I/O overhead that exceeds GPU computation time and leaves idle gaps between adjacent layers. To fill these gaps, expert-prefetching strategies \cite{adapmoe,fate,klotski,hybridmoe} have been widely studied, yet existing solutions still fall short of addressing I/O inefficiency, poor adaptability to legacy hardware, and lack of global scheduling, which motivates our proposed LayerScope system.
Fate \cite{fate} prefetches experts during attention computation but cannot resolve the fundamental mismatch between compute duration and I/O latency. Klotski \cite{klotski} increases batch size to overlap I/O, but at the cost of activating more experts. kTransformers \cite{ktransformers} pre-distributes experts across CPU and GPU for in-place computation, performing well in single-batch inference but suffering from CPU bottlenecks in multi-batch scenarios—especially on older CPUs with weaker single-core performance. Fiddler \cite{fiddler} dynamically places experts across CPU and GPU, yet lacks precise scheduling for hot experts (those processing more tokens). HybriMoE \cite{hybridmoe} introduces a queuing mechanism to prioritize hot experts on the GPU, but its greedy per-layer placement overlooks the cross-layer optimization opportunities enabled by prefetching.
A comparative analysis is provided in Table \ref{copal}, underscoring the need for a unified scheduling system that addresses I/O inefficiency, legacy hardware adaptability.

While CPU–GPU co-execution offers a promising direction, exist-While CPU–GPU co-execution offers a promising direction, existing solutions struggle with three fundamental challenges:
\textbf{Predic-}

\noindent \textbf{tion Uncertainty.} Expert activation varies sharply across layers. Traditional prediction methods based on statistics or gating inputs fail to track these shifts (detailed in \S\ref{32}), causing mispredictions that waste PCIe bandwidth (Case2 in Figure \ref{figure1}).
\textbf{PCIe Resource Competition.} Prefetch and on-demand loads share limited bandwidth, and the layer-by-layer execution model always prioritizes the latter. Without coordination, when conflicts arise, prefetch degrades to on-demand loading of the next layer.
\textbf{Cross-Device Scheduling Complexity.} CPUs and GPUs differ dramatically in throughput and latency, forcing prior work to schedule each expert greedily as soon as a thread is free. The inability to model prefetch gains and losses results in suboptimal performance.

\begin{figure}[t]
  \centering
  \begin{subfigure}[ht]{0.49\linewidth}
    \includegraphics[width=\linewidth]{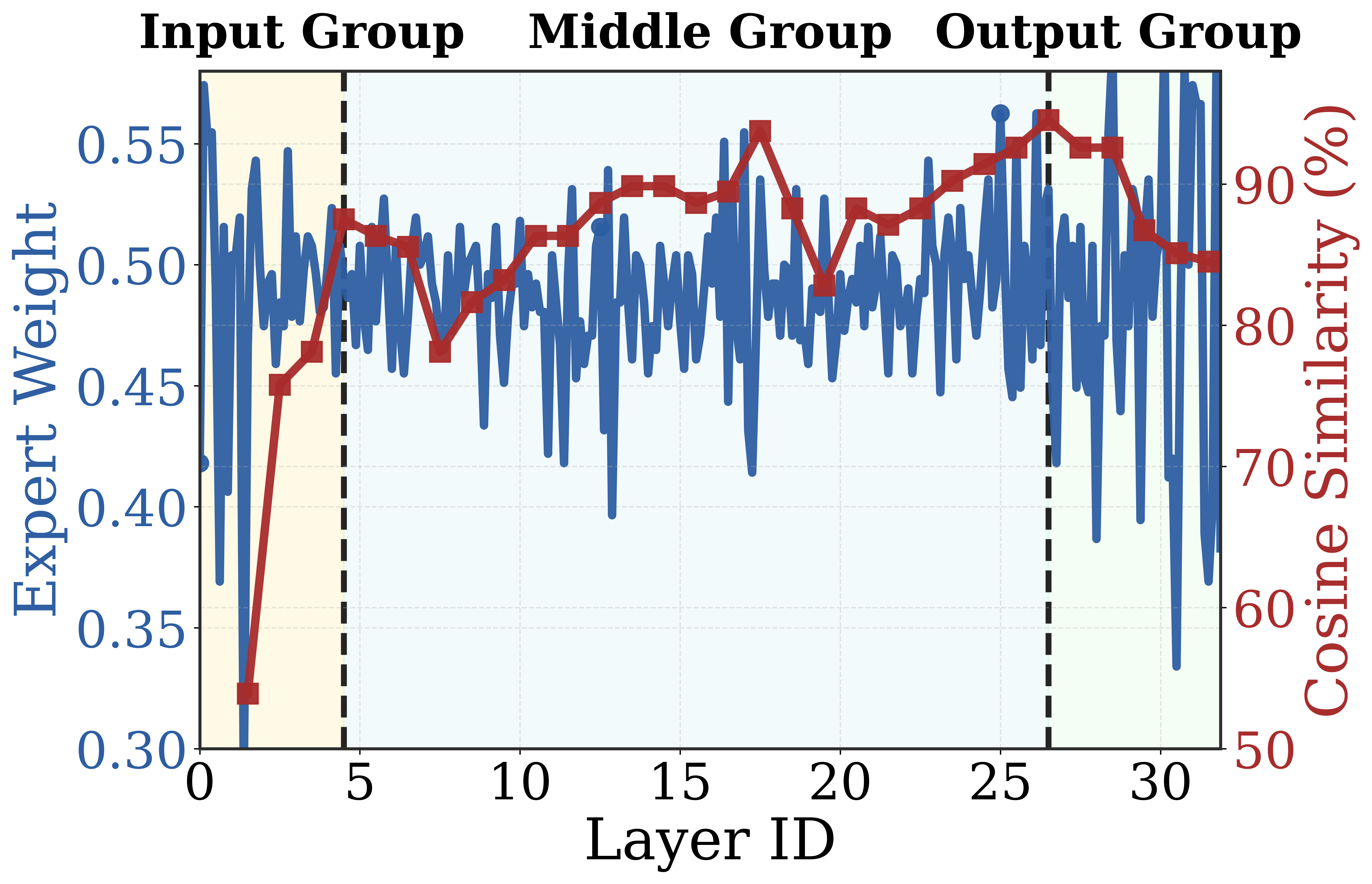}
  \end{subfigure}
  \begin{subfigure}[ht]{0.49\linewidth}
    \includegraphics[width=\linewidth]{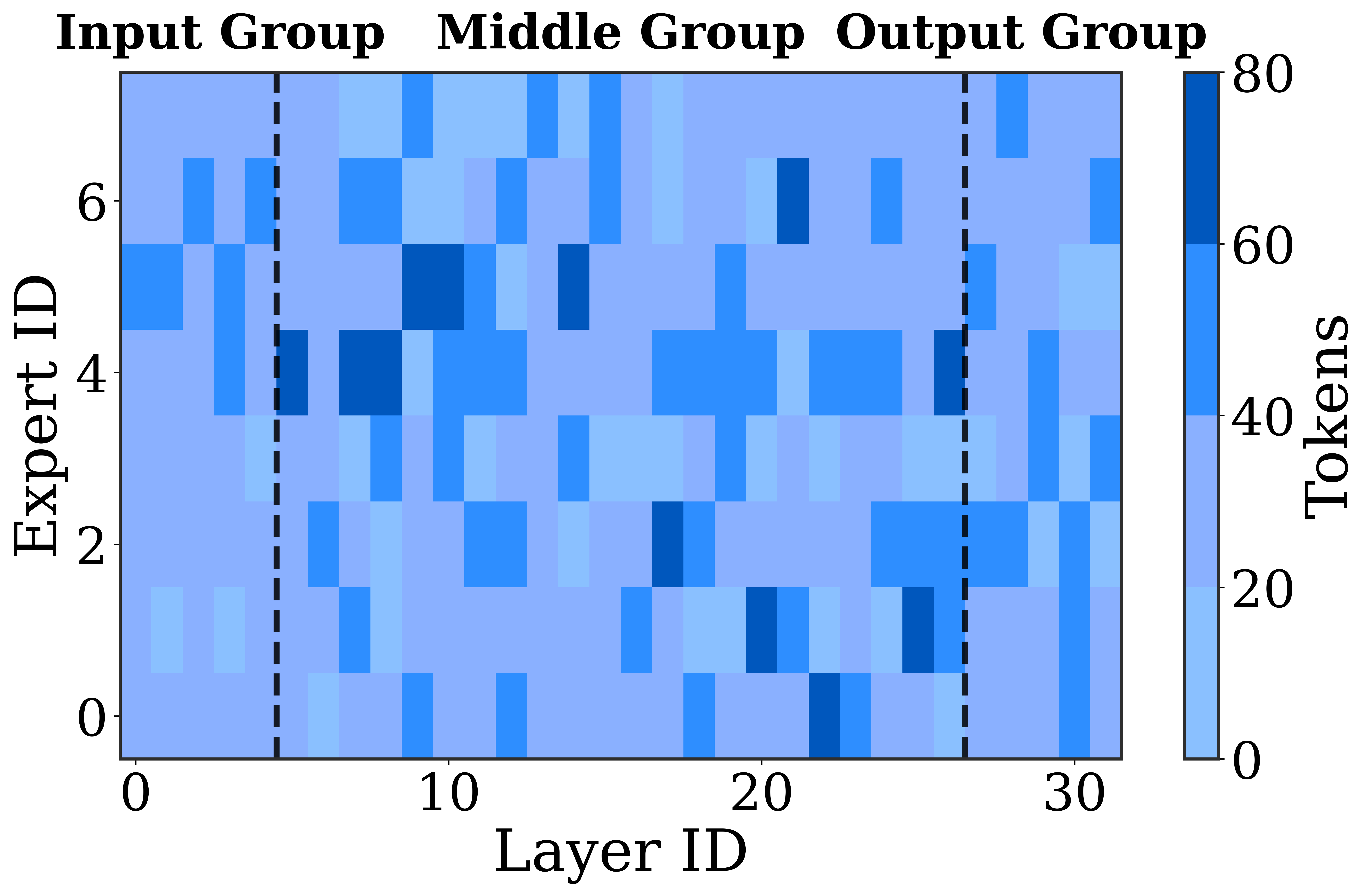}
  \end{subfigure}  
  \caption{Group-wise characteristics of expert weight, inter-layer cosine similarity, and activation distribution in MoE.}
     \label{figure2} 
\end{figure}

In response to these challenges, this paper proposes LayerScope, a prediction-driven cross-layer expert scheduling system. The core findings and contributions are as follows:

\textbf{1. Layer-group properties.} Earlier work observed inter-layer routing correlation \cite{exploiting}, cosine similarity of gating inputs \cite{hybridmoe,fate}, and hot-expert emergence \cite{toward,fiddler,prediction}. We show that these properties obey \emph{layer-group} structure: MoE networks naturally decompose into \textbf{input}, \textbf{output}, and \textbf{middle} groups.  As shown by Mixtral-8×7B in Figure~\ref{figure2}, input and output groups exhibit stronger routing correlation and more skewed gating weights, whereas middle layers show higher cosine similarity and dominant hot experts.

\textbf{2. Cross-layer prefetch efficiency.} Figure~\ref{figure1} (Case~3 vs.~4) shows that the layer-by-layer scheduling loads two experts to GPU to accelerate next layer startup, and repeats this at Layer 1, resulting in more I/O. Cross-layer scheduling instead uses one of the I/O slots to prefetch Layer~1 experts, slightly increasing Layer 0 latency yet reducing total latency. Passive, local prefetching inflates CPU idle time; globally, it fails to pipeline GPU, I/O, and CPU.  Prefetch must therefore be managed proactively inside a global cost model.

\textbf{3. Predictability of inference overhead.} CPU time grows linearly with token count, while PCIe and GPU times are largely token-invariant.  By acquiring the token distribution of the current and prefetch-target layers, we can estimate cross-layer latency via pipeline modelling.

\begin{table}[t!]
\caption{Comparison of MoE offloading systems.}
\centering
\resizebox{\linewidth}{!}{%
\begin{tabular}{c c c c c}
  \toprule
  Work & \makecell{Expert\\Computation} & \makecell{Prefetch\\Strategy} & \makecell{Scheduling\\Strategy} & \makecell{Load\\Balancing} \\
  \midrule

  Fate \cite{fate} & GPU   & \makecell{Gate}  & \makecell{On-Demand}       & \XSolidBrush \\
  Klotski \cite{klotski}           & GPU                    & \makecell{Expert}   & \makecell{On-Demand}       & \ding {73} \\
  Fiddler \cite{fiddler}           & CPU+GPU                & None                 & \makecell{Layer-by-Layer}       & \ding {73} \\
  HybriMoE \cite{hybridmoe}          & CPU+GPU                & \makecell{Gate}     & \makecell{Layer-by-Layer}       &\ding {72} \\
  \hline
  ours           & CPU+GPU                & \makecell{Expert-Gate} & \makecell{Cross-Layer} & \ding {72}\ding {72} \\
  \bottomrule
\end{tabular}%
}
\label{copal}
\end{table}

Based on these insights, LayerScope introduces three co-designed innovations:

\begin{itemize}[leftmargin=0.05cm,itemindent=0.5cm]
 
\item \textbf{LLaPor}: a layer-aware lightweight predictor that uses distinct architectures for each layer group, exploits both inter-layer routing correlation and adjacent-layer cosine similarity statistics, achieves a Top-4 prediction accuracy of over 90\%, with support for online continuous learning.

\item \textbf{PreSched}: a prefetch-aware cross-layer scheduler that builds a unified cost model to dynamically optimize expert placement by quantifying the global benefits of prefetching and on-demand loading, thereby breaking through the limitations of greedy strategies.

\item \textbf{AsyncIO}: an asynchronous I/O optimizer that decouples expert movement from computation, pipelines transfers across heterogeneous memories, and splits each expert into fine-grained chunks to saturate PCIe bandwidth.
\end{itemize}
Extensive evaluation on Mixtral-8×7B \cite{moe}, DeepSeek-MoE \cite{deepseekv2}, Qwen3-30B-A3B \cite{Qwen3}, and Moonlight-16B-A3B \cite{moonlight} shows that LayerScope boosts end-to-end inference throughput by 141\% over Klotski and 70.8\% over HybriMoE, while cutting decoding latency by 74.6\% and 42.1\%, respectively, providing an efficient solution for MoE deployment in resource-constrained environments.






\section{Background and Related Work}
\label{II}
\subsection{Inference of MoE Models}
\label{21}

MoE is one of the dominant architectures for large language models \cite{gpt4,deepseekv2,moe}. A MoE model consists of multiple stacked MoE blocks. For an input token $x$, the processing within one MoE block can be summarized as follows: the hidden state $a$ is computed by the attention module and fed into a gating network, which uses a softmax to compute routing weights $w_i$ and selects the top-\(k\)experts ${e}_i$. Their outputs are aggregated via a weighted sum to produce the block output $x'$, which is then passed to the next block. This process can be be expressed as:
\[
a = \text{Attention}(x) \quad e_i, w_i = \text{Gate}(a) \quad x' = \sum_{i=0}^{k-1} w_i \cdot \text{Expert}_i(a)
\]

A key observation is that once the hidden state $a$ and current gating output (${e}_i$, $w_i$) are fixed, the expert activation pattern in subsequent layers becomes deterministic, since the parameters of all functional modules remain unchanged.


\begin{figure}[t]
  \centering
  \begin{subfigure}[ht]{\linewidth}
    \includegraphics[width=\linewidth]{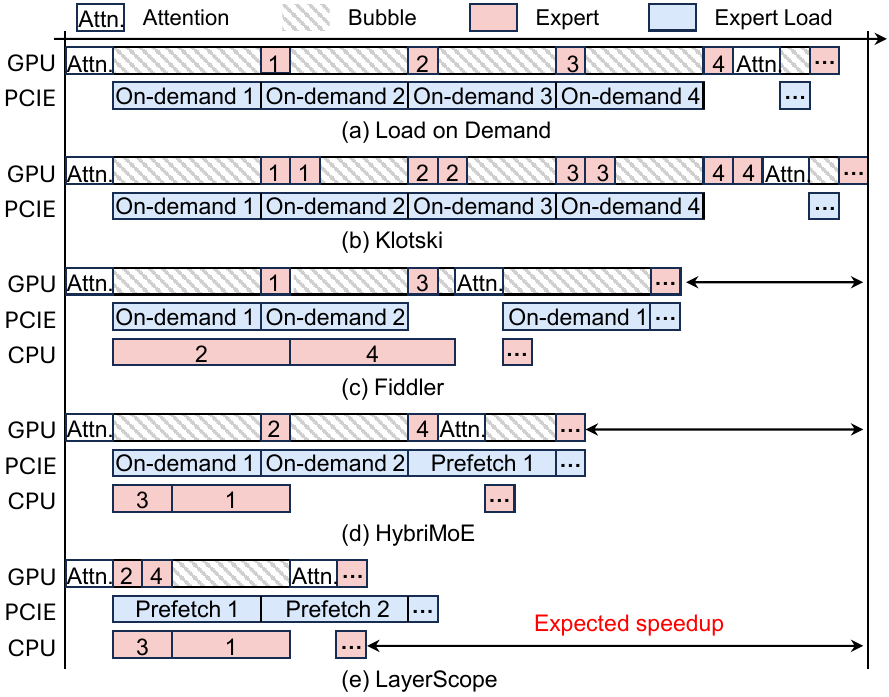}
  \end{subfigure}
  \caption{Execution timeline: LayerScope shrinks bubbles via accurate prefetch and cross-layer scheduling.}
     \label{figure4} 
\end{figure}
\subsection{MoE Offloading}


To overcome the memory constraints~\cite{specexec,dovetail,Moelightning,PowerInfer} of large language models, offloading expert parameters to host memory has become a necessary practice. This, in turn, makes the expert-placement problem a central determinant of system performance. 

On-demand loading methods, such as Lina~\cite{lina} and ExpertFlow~\cite{expertflow}, exclusively utilize GPUs for expert computation. If an expert is not prefetched to the GPU, it must be loaded on demand, as shown in Figure~\ref{figure4}. The downside of these approaches is the high I/O overhead, which causes prolonged GPU idling. MoE-GEN~\cite{MoE-GEN} mitigates this by deferring expert loading until the number of tokens targeting that expert reaches a preset threshold. Likewise, Klotski~\cite{klotski} enlarges the batch size to amortize the cost of loaded experts, thereby masking I/O overhead to some extent. To further reduce the I/O bubble, kTransformers~\cite{ktransformers} pre-distributes experts across the CPU and GPU, with each expert performing computation in place without movement; this approach is suitable for single-batch inference but faces CPU bottlenecks in multi-batch scenarios. In contrast, Fiddler~\cite{fiddler} allows expert computation on the CPU and leverages on-demand expert loading to improve GPU utilization, but it struggles to accurately offload high-workload experts to the GPU. 
To address this issue, HybriMoE~\cite{hybridmoe} introduces a queuing mechanism that proactively offloads cold experts to the CPU, aiming to increase the reuse probability of hot experts on the GPU. This strategy, however, fails in memory-constrained environments: without sufficient GPU memory to guarantee the reuse of retained experts, the proactive offloading merely adds extra I/O overhead. Furthermore, the design overlooks the critical conflicts when prefetching coexists with on-demand loading and unloading—namely, contention for limited PCIe bandwidth and expert residency conflicts under strict capacity constraints.

\subsection{CPU-GPU Collaborative Inference}

\begin{figure}[t]
  \centering
  \begin{subfigure}[ht]{\linewidth}
    \includegraphics[width=\linewidth]{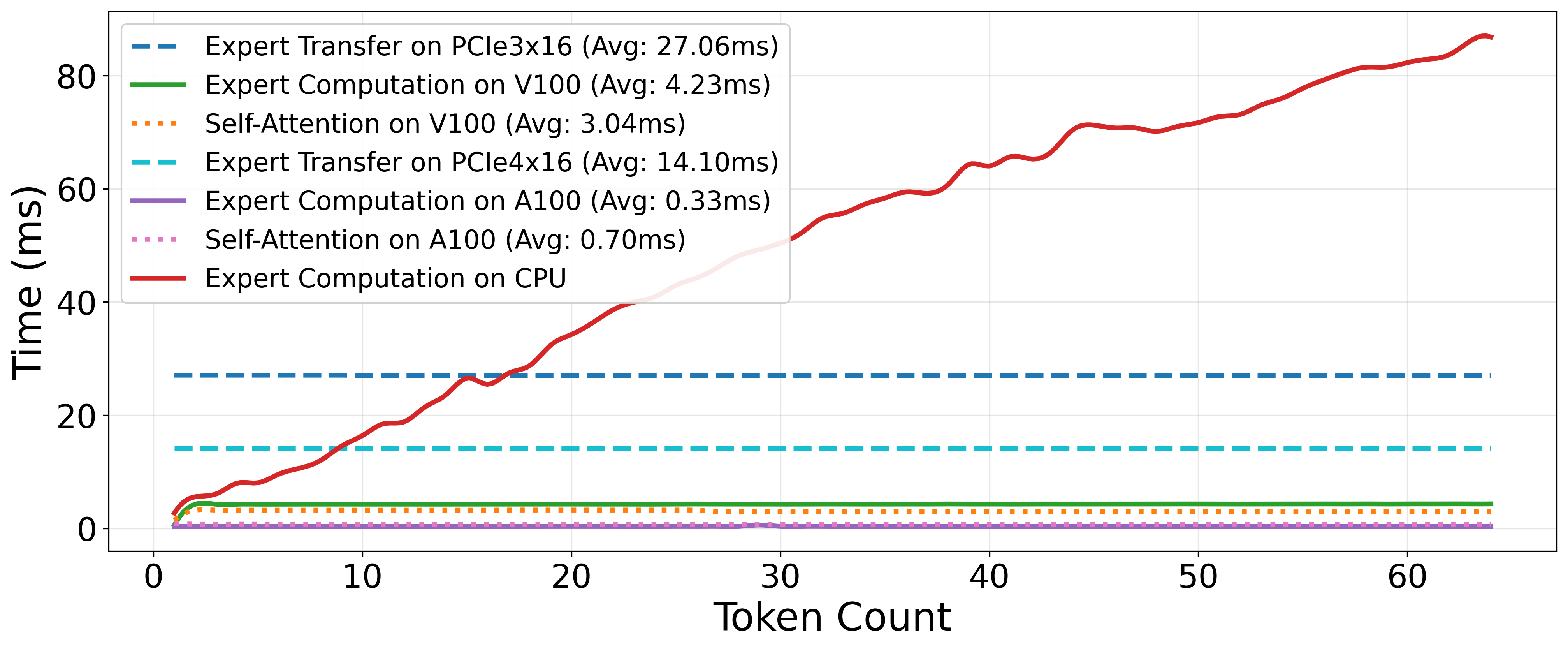}
  \end{subfigure}
  \caption{Characterization of expert computation and transfer costs across heterogeneous devices.}
     \label{figure5} 
\end{figure}

Improving cooperative-inference efficiency~\cite{aptmoe,flexinfer,hetegen} is a system-level challenge that embraces model compression~\cite{lut,MoQAE,qmoe,MxMoE}, 

\noindent activating fewer experts~\cite{Lynx,daop,dynamic,lu2024not}, and kernel-level optimizations~\cite{liquidgemm,pagedattention,flashattention,ktransformers}.  
Our goal is orthogonal: we optimize without sacrificing accuracy and remain model-agnostic and hardware-portable. 
Figure~\ref{figure5} quantifies the token-count dependence of compute and I/O cost for Mixtral-8×7B.  
Because the CPU’s parallel throughput is limited, its per-expert compute cost grows almost linearly with the number of tokens, whereas GPU compute and PCIe-transfer costs stay flat.  
With optimizations such as pinned host memory and asynchronous transfers, we achieved 79\% PCIe bandwidth utilization. Under these conditions, transferring two experts concurrently costs no less than transferring them serially, so expert traffic can be modelled as a sequential process. These observations form the basis of our inference cost model.

\subsection{MoE Prefetching}

In MoE inference, prefetching experts for subsequent layers increases I/O–computation overlap~\cite{fMoE,hybridmoe,fate}. Prior methods exploit distinct activation patterns: Klotski~\cite{klotski} and MoE-Infinity~\cite{MoE-Infinity} rely on the gating network’s bias toward hot experts; ExFlow~\cite{exploiting} detects inter-layer routing correlations; Fate~\cite{fate}, InfiniGen~\cite{InfiniGen} and HybriMoE~\cite{hybridmoe} use the cosine similarity of gating inputs to predict expert activation. Another line of research adopts learned predictors: ExpertFlow~\cite{expertflow} and OpenMoE~\cite{OpenMoE} match token-level similarity to forecast activations, whereas SiDA~\cite{sida} applies embedding-based hashing and exceeds 90\% hit rate on Switch-Base-128~\cite{switch}. Pre-Gated MoE~\cite{Pre-gated} and ProMoE~\cite{ProMoE} employ explicit modeling of gating-activation correlations to implement a uniform predictor.

However, analysis across Mixtral~\cite{moe}, DeepSeek~\cite{deepseekv2}, Moonlight~\cite{moonlight}, and Qwen3~\cite{Qwen3} shows that these properties exhibit pronounced layer-group dependence rather than uniform behaviour.  A single heuristic therefore gives only limited accuracy within specific layer groups; effective prediction requires combining multiple features tailored to the target group, as we elaborate in \S\ref{31}.


\section{Motivation}

\subsection{Observation of Expert Layer Groups}
\label{31}

As a case study in Figure \ref{figure2}, Mixtral-8x7B illustrates the correlation between expert routing weights, input cosine similarity, expert distribution, and layer groups. The same pattern holds for the other models. Building on \S\ref{21}, we further analyze: in input and output layer groups, certain expert weights \( w_i \) dominate the output aggregation, amplifying the correlation of expert routing across layers. Hence, predictors should weigh previous-layer activations and their routing weights more heavily. In middle layer groups, balanced expert weights lead to the hidden state \( a \)  highly similar across layers, and the router tends to select hot experts. Thus, predictors here should focus more on the hidden state \( a \) from the preceding layer.

\begin{figure}[t!]
  \centering
  \begin{subfigure}[ht]{\linewidth}
    \includegraphics[width=\linewidth]{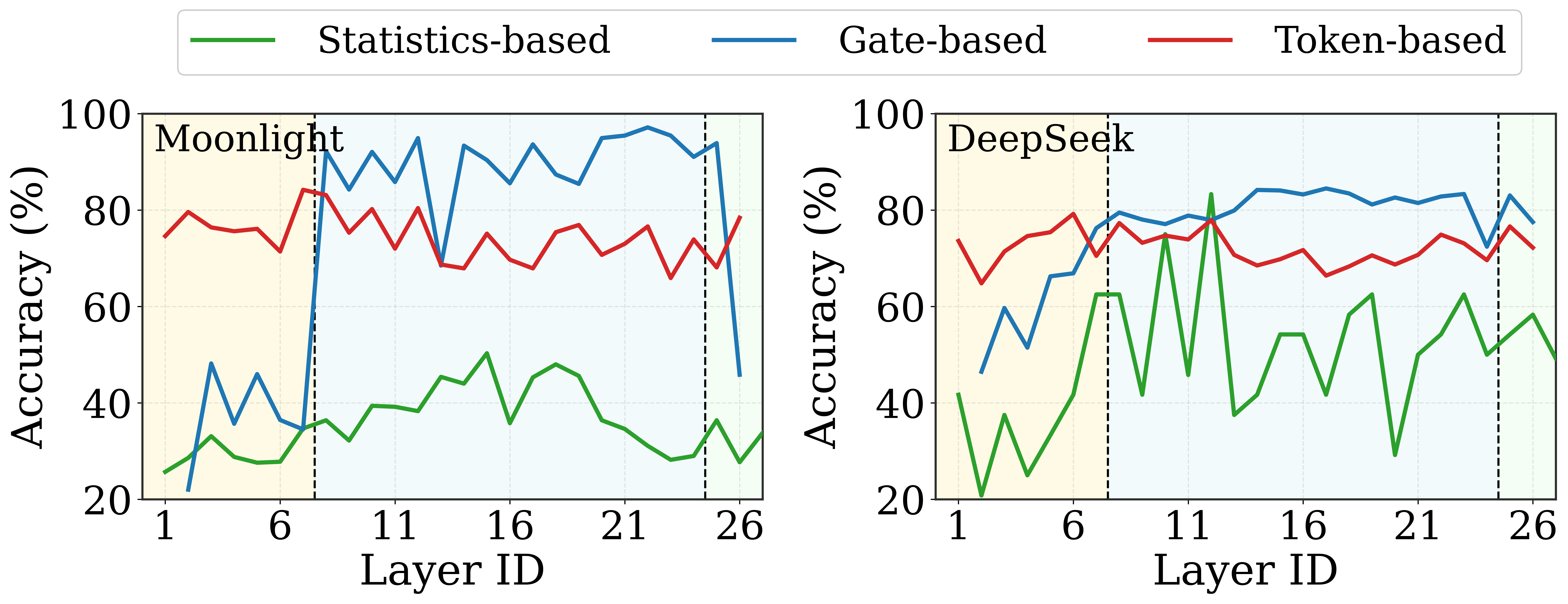}
  \end{subfigure}
  \caption{Comparison of three existing expert activation prediction strategies.}
     \label{figure6} 
\end{figure}

\subsection{Limitations of Existing Prediction Methods}
\label{32}
Figure \ref{figure6} compares the expert prediction accuracy of three strategies during inference decoding in DeepSeek and Moonlight. The statistics-based method is only effective in middle layer groups by leveraging hot expert distributions. The gate-based strategy, which utilizes high cosine similarity of gating inputs across adjacent layers, performs especially well in middle layer groups, with slightly lower accuracy in input and output groups. The token-based approach employs a light predictor to model token-expert and expert-expert correlations, achieving better performance in input and output layer groups.
These comparisons highlight that leveraging expert activation properties must be layer-group-aware. Manually identifying or statistically capturing these properties per group remains challenging. In contrast, using relevant variables as input features for a neural predictor to learn these associations in a data-driven manner offers a more feasible solution.

\begin{figure}[t!]
  \centering
  \begin{subfigure}[ht]{\linewidth}
    \includegraphics[width=\linewidth]{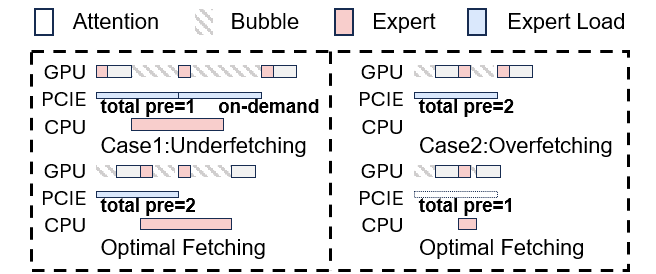}
  \end{subfigure}
  \caption{Two suboptimal prefetching scenarios caused by layer-by-layer scheduling.}
     \label{figure7} 
\end{figure}

\subsection{Limitations of Layer-by-Layer Scheduling}

The mainstream approach to expert placement in CPU–GPU cooperative inference uses a greedy, layer-by-layer scheduler that balances CPU cost $T_C$ and GPU cost $T_G$ to minimize the current layer’s completion time \cite{hybridmoe}. This is formulated as:

\begin{equation}
\arg \min_{\substack{E_{cpu}, E_{gpu}}} \max(T_C, T_G)
\label{eq2}
\end{equation}

When on-demand loading coexists with prefetching, however, we observe two systematic inefficiencies. First, if the previous layer is GPU-bound ($T_G$ > $T_C$), its on-demand loads preempt prefetch bandwidth for the target layer, forcing additional on-demand loads later and raising end-to-end latency (Figure~\ref{figure7}, Case~1). Second, if the previous layer is CPU-bound ($T_C$ > $T_G$), the scheduler greedily fills the PCIe pipe by prefetching experts that should be on the CPU, delaying the target layer’s compute start and again increasing latency (Figure~\ref{figure7}, Case~2).

The root cause is that layer-wise schedulers optimise only the current layer’s cost, ignoring the activation pattern of the prefetch-target layer. This results in uncertain prefetching decisions—whether and how many experts to prefetch. Therefore, incorporating the expert activation status of both the current and target layers into the scheduling decision space can enable more efficient prefetch scheduling in terms of timing and frequency.

\section{LayerScope Design}
\subsection{Overview}

\begin{figure}[t]
  \centering
  \begin{subfigure}[ht]{\linewidth}
    \includegraphics[width=\linewidth]{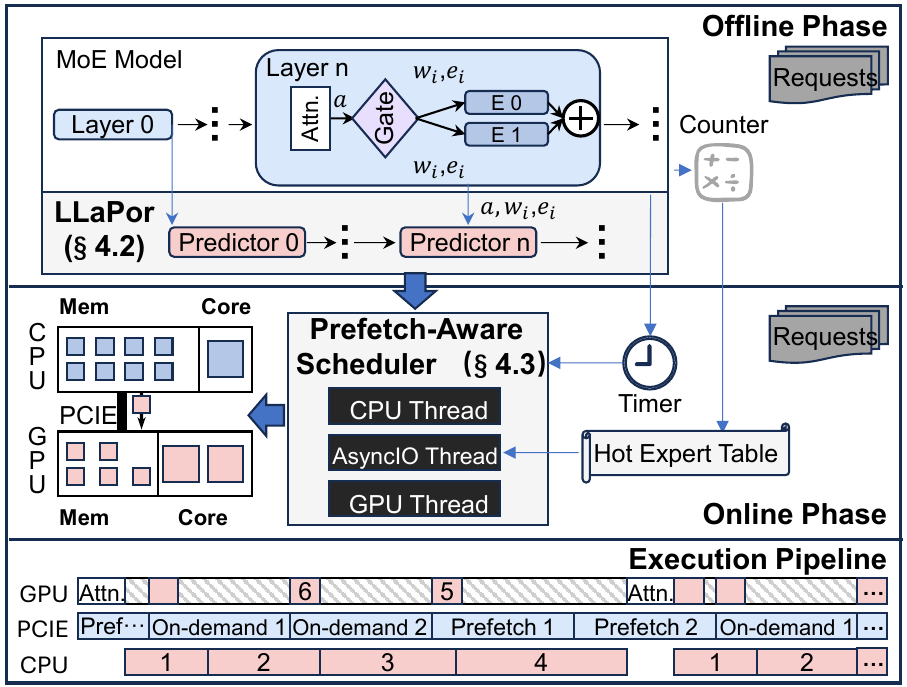}
  \end{subfigure}
  \caption{System overview of LayerScope.}
     \label{figure8} 
\end{figure}

We present LayerScope, a predictive cross-layer scheduling system that enables efficient multi-batch MoE inference on legacy servers by holistically tackling the dual bottleneck of memory capacity and PCIe latency. Figure~\ref{figure8} shows the system overview of LayerScope.  

In the offline phase we profile the target model over multiple iterations to build the dataset for training the Learnable Layer-Aware lightweight Predictor (LLaPor). The Timer records per-operation costs, while the Counter ranks experts by the number of processed tokens to produce a hot-expert table. Given the available GPU memory, the system loads the hottest experts in descending order until the memory budget is exhausted.
During online inference the Timer continuously monitors the execution pipeline, while LLaPor predicts expert activations for upcoming layers from the current-layer inputs. The Prefetch-aware Cross-layer Scheduler (PreSched) combines Timer feedback with LLaPor forecasts to make real-time decisions that minimize pipeline bubbles. The Asynchronous I/O Optimiser (AsyncIO) decouples expert movement from computation, transfers experts asynchronously across heterogeneous memories, and splits each transfer into fine-grained chunks to maximize PCIe throughput.
LLaPor continually ingests new runtime data to adapt to evolving task patterns.

\subsection{Learnable Layer-aware Predictor}

\subsubsection{Design principle}
According to \S\ref{21}, there exists a deterministic functional mapping from the tuple ($a$, ${e}_i$, $w_i$) of the current layer to the set of activated experts ${e}_i'$ of the next layer. Consequently, expert activation patterns are intrinsically predictable given the preceding hidden state and routing decisions. This predictability forms the theoretical foundation for designing LLaPor.

\subsubsection{Network Architecture of LLaPor}

\begin{figure}[t]
  \centering
  \begin{subfigure}[ht]{\linewidth}
    \includegraphics[width=\linewidth]{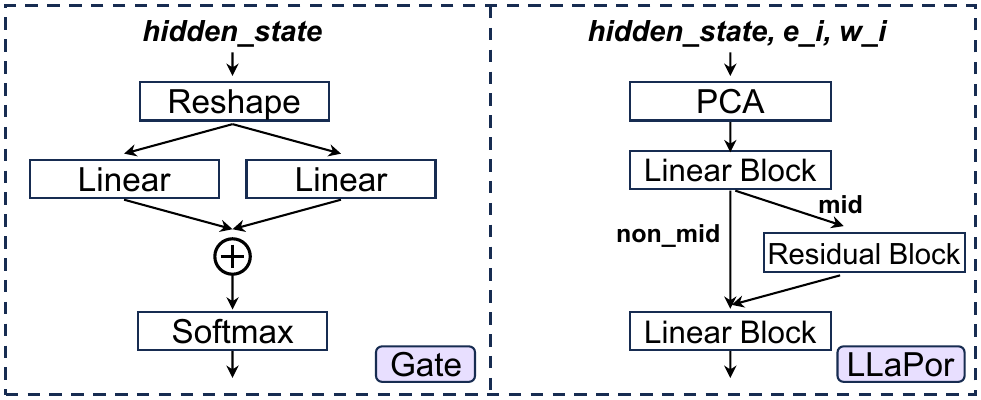}
  \end{subfigure}
  \caption{Architectural comparison between the standard gating network and LLaPor.}
     \label{figure9} 
\end{figure}

LLaPor is inspired by the native MoE gating network but builds an independent lightweight predictor for every MoE block. As shown in Figure~\ref{figure9}, LLaPor first compresses the hidden state \( a \) with Principal Component Analysis (PCA) to cut training cost, then fuses the reduced features. Instead of the single linear transform used in the gating network, each linear block in LLaPor consists of a linear layer, GELU activation, and Dropout for more robust feature learning.
Reflecting the layer-group differentiation in MoE models, predictors assigned to the input and output groups use shallow networks to avoid over-fitting on the noisier features typical of these layers. Predictors for the middle groups are deeper and contain an additional residual block composed of stacked linear layers, GELU activations and dropout; a gating unit modulates this block to capture the influence of the hidden state \( a \) on expert activation.

\subsubsection{Training Strategy of LLaPor}

LLaPor employs a layer-group-differentiated training strategy, with variations in the loss function, feature variables, and optimizer.

\noindent\textbf{Dataset.}
We perform offline inference on 128 batches from the ShareGPT dataset~\cite{sharegptv3}, each consisting 512 tokens. During inference we collect per-layer records ($a$, ${e}_i$, $w_i$). These records are organized by layer to form the training dataset. Cross-dataset evaluations that demonstrate generalization are presented in \S\ref{preacc}.

\noindent\textbf{Loss Function} Predictors in input and output groups are trained using a hybrid loss function designed to address class imbalance and improve learning on hard samples. The overall loss combines an expert-balancing loss \( L_{\text{expert}} \) and a focal loss term \( L_{\text{focal}} \), expressed as \( L = L_{\text{expert}} + \lambda \cdot L_{\text{focal}} \), where \( \lambda \) is a weighting hyperparameter. The expert-balancing term incorporates inverse expert frequency \( 1/f_i \) to mitigate activation imbalance, and is defined as:
\[
L_{\text{expert}} = -\frac{1}{N} \sum_{i=1}^{N} \frac{ \text{BCE}(y_i, p_i)}{f_i}
\]
where \( y_i \in \{0,1\} \) indicates whether expert \( i \) is activated, \( p_i \) denotes the predicted probability of expert \( i \), \( N \) is the total number of experts, and BCE refers to the Binary Cross-Entropy loss \( \text{BCE}(y_i, p_i) = y_i \log(p_i) + (1 - y_i) \log(1 - p_i) \). The focal loss term emphasizes hard misclassified examples by reducing the contribution of easy samples through a modulating factor \( (1 - p_t)^\gamma \), where \( p_t \) is the predicted probability for the true label and \( \gamma \geq 0 \) is a focusing parameter. This term is given by:
\[
L_{\text{focal}} = -\frac{1}{N} \sum_{i=1}^{N} (1 - p_t)^\gamma \cdot \text{BCE}(y_i, p_i)
\]

\begin{figure}[t!]
  \centering
  \begin{subfigure}[ht]{\linewidth}
    \includegraphics[width=\linewidth]{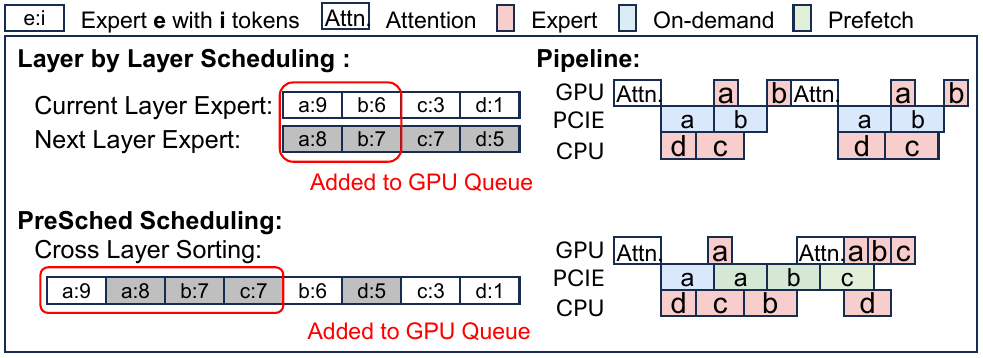}
  \end{subfigure}
  \caption{Comparison of PreSched scheduling versus layer-by-layer scheduling strategies.}
     \label{figure10} 
\end{figure}

\noindent\textbf{Feature Variables}
Strong PCA compression (4096-dim to 256-dim) is applied to hidden state \( a \) in input and output groups due to high cross-layer variation; middle groups use milder compression (4096-dim to 512-dim) thanks to feature stability.

\noindent\textbf{Optimizer.} AdamW is used with learning rates and weight decay dynamically adjusted by layer. Input and output groups employ a composite scheduler: linear warmup for 5 epochs followed by cosine annealing for faster convergence. The middle group uses higher learning rates and weight decay to accelerate convergence and counter overfitting, leveraging their stable features.

\noindent\textbf{Online fine-tuning.}
The initial offline training of LLaPor requires 30 epochs, each taking 1.2 seconds on a single A100 GPU. To maintain accuracy under shifting input distributions, LLaPor employs an online learning mechanism that monitors prediction error over a sliding window of the most recent 64 tokens in the input group. Once the average error exceeds 10\%, a lightweight fine-tuning cycle is triggered, using runtime variables collected from 128 tokens. Each online fine-tuning iteration completes in just 1.8 ms.

\subsection{Prefetch-Aware Cross-Layer Scheduling}

\subsubsection{Design Principle}

PreSched jointly optimizes expert placement for the current layer and the prefetch-target layer, maximizing the global benefit of every transfer instead of merely minimizing the current-layer latency. Traditional schedulers, confined to the current layer’s activation distribution, apply a greedy layer-wise policy. As Figure~\ref{figure10} illustrates, PreSched exploits LLaPor’s inter-layer predictions to expand the scheduling horizon to both layers, gaining cross-layer visibility. Under limited PCIe bandwidth it prioritizes the most heavily loaded experts, moving more tokens per transfer and shrinking PCIe idle windows. PreSched also calibrates the prefetch count to fully exploit CPU-compute bubbles for loading, while preventing excessive prefetch traffic from interfering with critical I/O.

\subsubsection{Problem Modeling}
\label{secmodel}

To model the problem, the following premises and constraints are proposed:

\noindent\textbf{Premise 1 (Cross-layer Sorting).} The gating network provides activated experts for the current layer, and LLaPor predicts those for the next layer. After excluding experts already on the GPU or being loaded, lists $E_{cur}[0\ldots n]$ and $E_{next}[0\ldots n'-1]$ are formed by sorting experts in ascending order of pending tokens. The two lists are then merged into $E_{all}[0\ldots n+n']$ via cross-layer ordering.

\noindent\textbf{Premise 2 (On-demand Monotonicity).} To maximize on-demand benefit, if $E_{cur}[i]$ requires no loading, then neither do $E_{cur}[0\ldots i-1]$; conversely, if $E_{cur}[i]$ requires loading, so must $E_{cur}[i+1\ldots n]$.

\textbf{Constraints.} Expert loading time is fixed as $t_{io}$; GPU expert computation cost is $t_g$; attention computation time is $t_{attn}$; CPU expert cost $t_c$ depends on token count $m$:
\begin{equation}
 t_c = \beta \cdot m + C     
 \label{equation6}
\end{equation}
where $\beta$ is the CPU computational cost of a single token, and $C $ is the startup cost.

\noindent\textbf{Premise 3 (Prefetch Non-interruptibility).} A prefetch, once started but not finished before the current layer’s computation ends, must be completed without initiating new prefetches to avoid blocking subsequent I/O.

\noindent\textbf{Premise 4 (Serial I/O).} PCIe transfers are executed sequentially\footnote{This reflects the typical system setting. Single-expert transfers already reach 66-80\% PCIe peak (Table~\ref{iotab4}); adding parallelism risks congestion without faster arrival.}.

\noindent\textbf{Premise 5 (Computation Masking).} The GPU computation cost of any expert that has been preloaded can be completely masked by the I/O bubble and is omitted from the latency model.

The decision variable is whether $E_{cur}[i]$ needs to be loaded onto the GPU, with $i$ recursively decremented from $n$ to 0 until no loading is needed. The objective is to minimize the total layer latency $\Sigma layer\_latency$.

\subsubsection{Trade-Off between On-Demand and Prefetch }

\begin{figure}[t]
  \centering
  \begin{subfigure}[ht]{0.8\linewidth}
    \includegraphics[width=\linewidth]{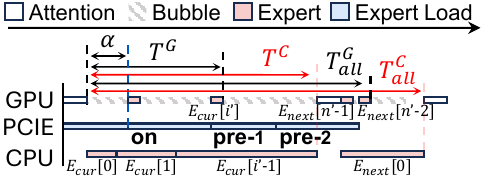}
  \end{subfigure}
  \caption{Mathematical modeling of PreSched.}
     \label{figure11} 
\end{figure}

The optimal placement of experts depends on the computation location of hot experts. Based on the problem modeling constraints, the issue simplifies to choosing for expert $E_{cur}[i]$ between strategy A (on-demand load to GPU) or B (compute directly on CPU, reserving I/O bandwidth for prefetching the next layer’s experts). To quantify this trade-off, we formulate four key questions, with Figure \ref{figure11} providing a visual summary of the relevant variables for clarity.

\noindent\textbf{Q1: Which experts must be moved to the GPU? }
We establish a cross-layer cost model for $E_{all}[i]$: transfer is mandatory if the total GPU cost $T^G_{all}$ is lower than the CPU cost $T^C_{all}$.
\begin{equation}
\begin{aligned}
T^G_{all}
&= \alpha + (n+n'-i+1)\cdot t_{io} + t_g\\
T^C_{all}
&= t_c(E_{all}[0\ldots i]) + t_{attn}
\end{aligned}
\end{equation}
Here, $\alpha$ is the delay in starting the on-demand operation due to previous prefetching, and ${t_c}(E_{all}[0\ldots i])$ is the total CPU computational cost of experts $E_{all}[0\ldots i]$, which can be calculated using Equation \ref{equation6}. If $T^G_{all} < T^C_{all}$, add $E_{all}[i\ldots n+n']$ to the GPU Queue.

Since the current and predicted layers may have misaligned loads, the subsequence $E_{cur}[i'\ldots n]$ belonging to the current layer in the GPU Queue is rechecked:
\begin{equation}
\begin{aligned}
T_G &= \alpha + (n-i'+1) \cdot t_{io} + t_g \\
T_C &= t_c(E_{cur}[0\ldots i'-1])
\end{aligned}
\label{localre}
\end{equation}
If $T_G<T_C$ holds, then $i'$ is the smallest index satisfying the condition; otherwise, the search continues inward.

\noindent\textbf{Q2: Which experts will be prefetch? }
The gap length released by the current layer's CPU computation is:
\begin{equation}
T_{gap}=t_c(E_{cur}[0\ldots i'-1]) - \alpha - (n-i'+1)\cdot t_{io} \nonumber
\end{equation}
The number of prefetches $|f|$ that can be overlapped is:
\begin{equation}
\begin{aligned}
f =  \frac{T_{gap}+t_{attn}}{t_{io}}
\end{aligned}
\end{equation}
and we round $f$ to the nearest integer $|f|$.  Experts $E_{{next}}[n'-|f|\ldots n'-1]$ are therefore selected for prefetching.

\noindent\textbf{Q3: How to quantify prefetch gain and loss?} Experts in the prefetch layer are priority-sorted, and the predictor is more accurate on hot experts; executing the highest-priority prefetch first therefore yields the largest payoff. The only prefetch that can delay an on-demand operation in the prefetch layer is the critical prefetch $|f|$.  
Its cost is $(|f|-f)\cdot t_e$ and its I/O benefit is $(f-|f|+1)\cdot t_e$.  
Let $R_{\text{hit}}$ and $R_{\text{miss}}$ be the hit and miss rates of the critical prefetch.  The overall benefit $\xi$ of the critical prefetch $|f|$ is calculated as follows:

\begin{algorithm}[t]
\caption{Prefetch-Aware Scheduling for Layer $l$ }
\label{alg:presched}
\SetAlgoVlined
\SetKwInOut{Global}{Initialization}
\SetKwComment{comment}{$\triangleright$\ }{}
\SetKwProg{Event}{Event}{}{end}
\SetKwFunction{OnCpuGpuDone}{OnCpuGpuDone}
\SetKwFunction{OnIoDone}{OnIoDone}
\SetKwFunction{OndemandLoop}{OndemandLoop}
\SetKwFunction{CrossLayerQueue}{CrossLayerQueue}
\SetKwFunction{PrefetchDecision}{PrefetchDecision}
\newcommand{\bluecomment}[1]{\textcolor{blue}{\Comment{#1}}}

\textbf{Initialization:}\\
\hspace*{0pt}$j=0$, $i=n$, $i'=n$  \bluecomment{current-layer window}\; 
\hspace*{0pt}$E_{{cur}}[0\ldots n]$, $E_{{next}}[0\ldots n'-1]$   \bluecomment{expert lists}\; 


\Event{\CrossLayerQueue{$E_{{cur}},E_{{next}}$}}{
    $E_{{all}}$ $\gets$ sort($E_{{cur}} \cup E_{{next}}$) by token count\;
    \For{$k=0$ \KwTo $n+n'$}{ 
        \If{${T^G_{all}}(k) < {T^C_{all}}(k)$}{
            add $E_{{all}}[k]$ into $\mathit{GPU\_Q}$; 
        }
    }
}
\Event{\OnIoDone{}}{
    \OndemandLoop{}\;
    \PrefetchDecision{}\;
}
\Event{\OnCpuGpuDone{}}{
    compute $E_{{cur}}[0\ldots i'-1]$ on CPU\;
    compute $E_{{cur}}[i'\ldots n]$ on GPU\;
}

\Fn{\OndemandLoop{}}{
    \ForEach{expert $e$ \textbf{in} $\mathit{GPU\_Q}$}{
            \If{$e$ in $E_{cur}$}{   
            $ i'\gets $the index $i'$ of $e$  \bluecomment{$e$=$E_{{cur}}[i']$}\;
            
            \If{${ T_G}(i') < {T_C}(i')$}{   
                $ i\gets i'$\;
                \textbf{break}\;
            }            
        }

    }
        \While{$i<n+1$  }{
            load $E_{{cur}}[i]$ \; 
            $i \gets i+1$ \bluecomment{load $E_{{cur}}[i'\ldots n]$ }\;
        }     
}
\Fn{\PrefetchDecision{}}{
    \If{$\mathit{GPU\_Q} \cap E_{next} = \emptyset$}{
        \Return\;
    }

    \If{$R_{{hit}}(f-|f|+1) > R_{{miss}}(|f|-f)$}{
    $ i\gets |f|$\;
    }
    \Else{
    $ i\gets |f|-1$\;
    }
        \While{$i>0$}{
            prefetch $E_{{next}}[n'-i]$ \;
            $i \gets i-1$ \bluecomment{prefetch $E_{{next}}[n'-i\ldots n'-1]$ }\;
        }    
}
\end{algorithm}

\begin{equation}
\xi = R_{hit} \cdot (f-|f|+1) \cdot t_{io} - R_{miss} \cdot (|f|-f) \cdot t_{io}
\end{equation}

If $\xi$ is greater than 0, prefetch $|f|$ can be regarded as a less costly on-demand operation; otherwise, prefetch $|f|$ degrades to an on-demand start in the next layer.

\noindent\textbf{Q4: When do prefetching yield no benefits?} 

When GPU Queue has no elements in $E_{{next}}$, it means that there is no GPU prefetching requirement (commonly seen in small batches). PreSched then extends the prediction window for cross-layer prefetching, following the same trade-off logic.

\subsubsection{Algorithm Design}

PreSched dynamically determines the computation location of each expert using a cross-layer cost model. The complete procedure is detailed in Algorithm \ref{alg:presched}. During the offline phase, it collects the variables described in \S\ref{secmodel}. In the online phase, it first places hot experts into the cross-layer GPU queue $GPU\_Q$ via cross-layer sorting. Then, the on-demand loading loop function re-examines $GPU\_Q$ to identify the smallest expert index $i'$ for the current layer that satisfies load balance, thereby determining the list of experts $E_{cur}[i'\ldots n]$ to be loaded on-demand to the GPU. In the prefetch decision function, PreSched weighs the hit–miss trade-off to decide which experts will be prefetched during CPU compute gaps. The entire process is event-driven. The \texttt{OnIoDone} event determines the experts to be loaded or prefetched to the GPU and completes the corresponding I/O transactions. The \texttt{OnCpuGpuDone} event manages expert computations on the CPU and GPU respectively, forming a dual compute–I/O pipeline.


%

\subsection{Asynchronous I/O Optimizer}

Unlike standard asynchronous transfer implementations that only decouple computation and I/O superficially, our design further optimizes cache management and fine-grained data partitioning to address PCIe bandwidth fragmentation and inter-layer prefetch conflicts—key pain points in legacy hardware deployments.

The core idea is to decouple computation from expert transmission, enhancing transfer efficiency through pipelined operations and increasing computation-I/O overlap, thereby significantly reducing inference latency. Specifically, the system maintains two types of pre-registered expert weight caches in GPU memory: one for on-demand loading, which uses dual buffer areas to alternate between expert computing and expert loading for immediate requests when no prefetching occurs; the other serves as a prefetch buffer, divided into two groups handling prefetch tasks for the current and next layers, respectively. 
\begin{table}[t!]
\centering
\small
\caption{Expert transfer performance and PCIe bandwidth utilization after AsyncIO optimization.}
\label{tab:asyncio_optimization}
\resizebox{\linewidth}{!}{%
\begin{tabular}{@{}lcccccc@{}}
\toprule
& \multicolumn{2}{c}{\textbf{DeepSeek}} & \multicolumn{2}{c}{\textbf{Qwen}} & \multicolumn{2}{c}{\textbf{Mixtral}} \\
\cmidrule(lr){2-3} \cmidrule(lr){4-5} \cmidrule(l){6-7}
 & Cost (ms) & BW Util. & Cost (ms) & BW Util. & Cost (ms) & BW Util. \\
\midrule
PCIe 3.0 ×16 & 1.39 & 76\% & 0.79 & 72\% & 26.7 & 80\% \\
PCIe 4.0 ×16 & 0.76 & 70\% & 0.44 & 66\% & 13.98 & 76\% \\
\bottomrule
\end{tabular}%
}
\label{iotab4}
\end{table}
This pre-registration of cache regions eliminates the overhead of repeated memory allocation and mapping for each transfer; subsequent expert movements involve only direct writes (overwrites) within these fixed buffers, substantially reducing setup overhead. The grouped rotation design ensures that prefetching stays one layer ahead of computation, providing sufficient time for expert transfer and avoiding expert overwrite issues caused by inter-layer prefetch conflicts, thus guaranteeing computational correctness and cache validity.
\begin{figure*}[t!]
  \centering
  \begin{subfigure}[ht]{\linewidth}
    \includegraphics[width=\linewidth]{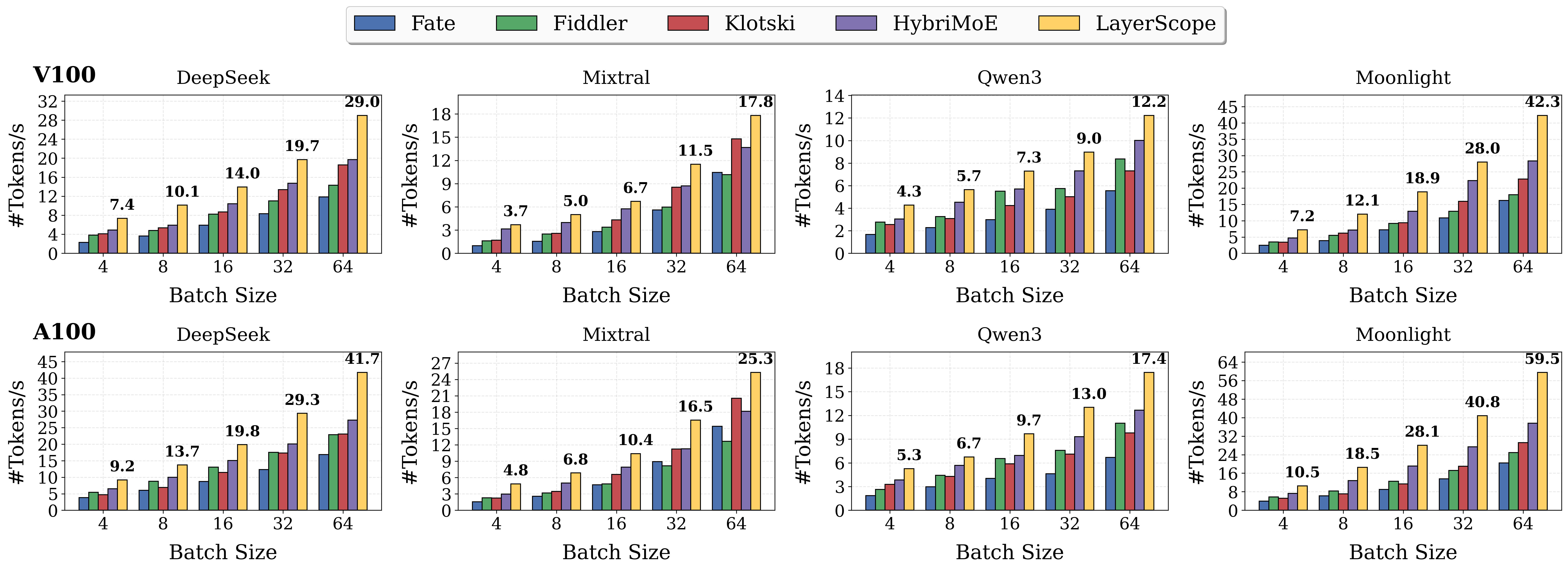}
  \end{subfigure}
  \caption{Throughput comparison between LayerScope and baseline systems across models and hardware.}
     \label{figure13} 
\end{figure*}

At the micro level each expert is decomposed into three linear weight matrices $gate\_proj$, $up\_proj$ and $down\_proj$.  Three independent CUDA streams are used to transfer these sub tensors concurrently.  This fine-grained strategy exploits the multi-lane capability of the PCIe bus by replacing one large transfer with several sub transfers that proceed in parallel. As a result, the scheduler gaps on the bus are filled, allowing throughput to approach the PCIe peak, as demonstrated by the high PCIe bandwidth utilization shown in Table \ref{iotab4}.

In addition all transfers use pinned host memory and asynchronous copy operations so that data movement proceeds without CPU intervention and does not block GPU computation. 
Compared to traditional synchronous loading, this mechanism achieves deeper computation-I/O overlap. By managing data movement within a fixed set of cache regions, it also guarantees correctness, thereby improving overall system throughput.

\section{Evaluation}

\subsection{Experimental Setup}

\noindent\textbf{Platform.}  
We evaluate on NVIDIA V100 and A100 GPUs. The V100 is connected via PCIe 3.0 x16, while the A100 uses PCIe 4.0 x16. Both are paired with an Intel Xeon Gold 6230R CPU. To facilitate a direct comparison and to model a memory-constrained environment (akin to consumer GPUs like the RTX 4090 with 24GB), we set a 24GB memory limit for both GPUs during our experiments.

\noindent\textbf{Models.}  
We select four widely adopted MoE models that exhibit diverse architectural characteristics: DeepSeek-V2-Lite \cite{deepseekv2}, Mixtral-

\noindent 8x7B \cite{moe}, Qwen3-30B-A3B \cite{Qwen3}, and Moonlight-16B-A3B \cite{moonlight}.
Table \ref{tab2} summarises the key differences. Mixtral represents models with fewer but larger experts, while Qwen3, Moonlight, and DeepSeek exemplify models with more but smaller experts.

\noindent\textbf{Baselines.}  
LayerScope is compared with four representative MoE inference frameworks\footnote{Omitted as KTransformers \cite{ktransformers} lacks pre-Ampere GPU support (e.g., V100) and is optimized for single-batch inference, underperforming in our multi-batch, resource-constrained scenario per HybriMoE \cite{hybridmoe}}: Fate \cite{fate} (GPU-based and highly accurate in prefetching), Fiddler \cite{fiddler} (CPU-GPU hybrid inference with on-demand loading), Klotski \cite{klotski} (reduces expert transfers by processing multiple batches in parallel), and HybriMoE \cite{hybridmoe} (CPU-GPU hybrid inference with greedy scheduling). Klotski limits itself to two micro-batch parallelism to avoid memory overflow. For a fair comparison, all baselines incorporate our optimized asynchronous expert-transfer method (utilizing pinned memory) and are initialized with our hot-expert table, while their native parameter-compression mechanisms are disabled.

\noindent\textbf{Metrics.}  
We measure throughput (tokens generated per unit of generation time) and decoding latency (time per output token). Generation time covers both the prefill and the decoding stages\footnote{This metric design aligns with Klotski~\cite{klotski}. }.All results are averaged over ten independent runs.

\begin{table}[t!]
\centering
\small
\caption{Configuration of evaluated MoE models.}
\label{tab:moe_config}
\resizebox{\linewidth}{!}{
\begin{tabular}{@{}lcccc@{}}
\toprule
 & \textbf{Mixtral} & \textbf{Qwen3} & \textbf{DeepSeek} & \textbf{Moonlight} \\
\midrule
Layers & 32 & 48 & 26 & 26 \\
Routed Experts & 8 & 128 & 64 & 64 \\
Activated Experts & 2 & 8 & 6 & 6 \\
Expert Memory (BF16) & 336 MB & 9 MB & 16.5 MB & 16.5 MB \\
\bottomrule
\end{tabular}%
}
\label{tab2}
\end{table}

\noindent\textbf{Dataset.}  
Experiments use the ShareGPT dataset \cite{sharegptv3} with batch sizes ranging from 4 to 64, an input length of 512 tokens, and an output length of 64 tokens. 
To verify the generalization capability of LLaPor, we additionally evaluate it on the instruction-following DPO dataset \cite{DPO} and the vision-language LLaVA dataset \cite{llava}.

\subsection{End-to-End Performance}


\subsubsection{Throughput}

Figure \ref{figure13} illustrates the end-to-end throughput of four models on V100 and A100 GPUs versus batch size. LayerScope consistently outperforms all baselines across configurations. Two scenarios are observed:

  


\begin{figure*}[t]
  \centering
  \begin{subfigure}[ht]{\linewidth}
    \includegraphics[width=\linewidth]{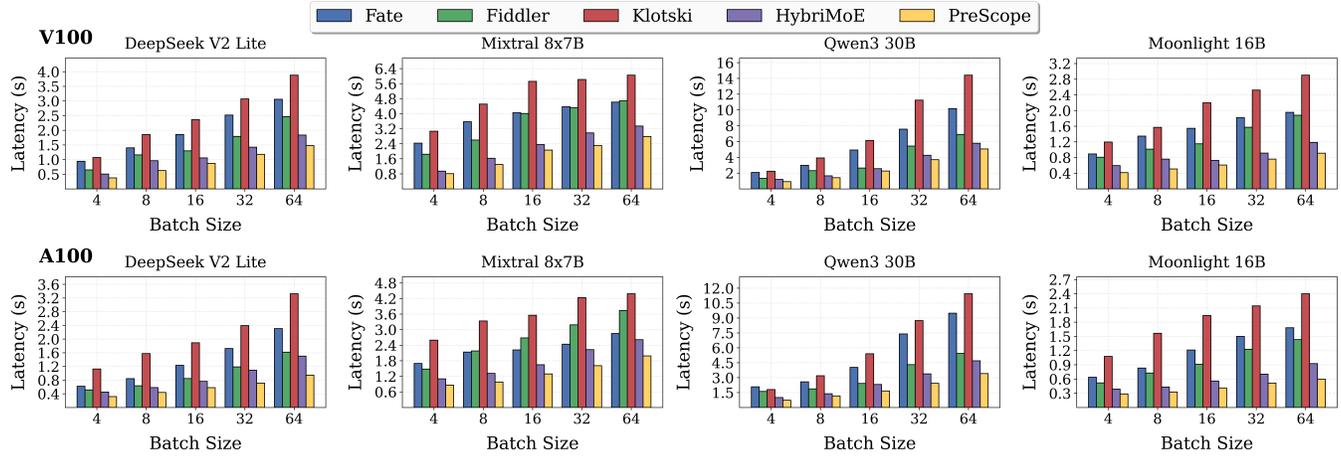}
  \end{subfigure}
  \caption{Decoding latency comparison between LayerScope and baseline systems across models and hardware.}
     \label{figure14} 
\end{figure*}

\textbf{Small-Expert Models.} These models contain numerous small experts per layer, making I/O latency the dominant bottleneck. GPU-only baselines such as Fate and Klotski exhibit significantly lower throughput due to frequent expert transfer overhead. When the batch size exceeds 8, Klotski’s multi-batch pipeline fails to achieve throughput scaling proportional to batch size, primarily because of the increased number of activated experts. On V100 with a batch size of 64, LayerScope achieves throughput improvements of 55.7\%, 67.2\%, and 85.6\% over Klotski, and 145\%, 120\%, and 161\% over Fate on DeepSeek, Qwen, and Moonlight, respectively. This advantage stems from LayerScope’s ability to intelligently offload part of the computation to the CPU, avoiding redundant expert transfers and thereby yielding significantly higher throughput.
On A100 with a batch size of 64, LayerScope achieves throughputs of 41.7, 17.4, and 59.5 tokens/s on DeepSeek, Qwen, and Moonlight, respectively. Compared to CPU-GPU collaborative methods, it surpasses HybriMoE by 52.7\%, 37.6\%, and 58.4\%, and outperforms Fiddler by 82.3\%, 59.6\%, and 138\%. This advantage stems from LayerScope's accurate expert prefetching mechanism and efficient scheduling optimization, which enable a higher degree of GPU-CPU parallelism. In contrast, HybriMoE, which employs a layer-by-layer scheduling strategy, shows limited performance at smaller batch sizes due to the negative I/O overhead introduced by its prefetching and dynamic expert offloading.

\textbf{Large-Expert Models.} The Mixtral model, with only 8 experts per layer but each occupying 336 MB, exhibits high computational intensity. As batch size increases, the CPU-GPU overhead gap widens, and LayerScope's margin over GPU-only baselines narrows. On V100, the speedup over Klotski and Fate decreases from 92.1\% and 219\% at batch size 8 to 14.1\% and 58.3\% at batch size 64. Nevertheless, LayerScope maintains a clear advantage over CPU-GPU collaborative baselines: at batch size 64 on A100, it exceeds HybriMoE by 39\% and Fiddler by 99\%. This result indicates that in compute-intensive scenarios, expert scheduling quality directly determines the pipeline bubble ratio. Fiddler's inefficient scheduling fails to adapt to batch size variations, while HybriMoE's greedy strategy, which prioritizes on-demand loading, contends with prefetch traffic, yielding limited gains. LayerScope explicitly quantifies the trade-off between prefetch benefits and compute offload, dynamically selects CPU or GPU execution for each expert, and compresses pipeline bubbles to the theoretical minimum.

Overall, across V100/A100, four models, and batch sizes from 4 to 64, LayerScope demonstrates excellent scalability, exhibiting the steepest performance growth curve in all tested scenarios. On the A100 platform, its advantage is more pronounced, particularly at larger batch sizes where precise prefetch control more efficiently leverages PCIe bandwidth enhancements for expert transfer gains. In summary, LayerScope improves throughput over Fate, Fiddler, Klotski, and HybriMoE by up to 264\%, 129\%, 141\%, and 70.8\%, respectively, scaling with PCIe bandwidth and establishing a new performance benchmark for MoE architectures.

\subsubsection{Decoding Latency}

Figure \ref{figure14} compares the average per-token decoding latency of LayerScope and four baselines on V100 and A100 GPUs. Across all batch sizes and model architectures, Layer-

\noindent Scope consistently achieves the lowest latency, owing to its precise compression of the bottleneck path through accurate prediction and cross-layer scheduling.

\textbf{Small-Expert Models:} On V100 at batch size 4, LayerScope achieves 0.37 seconds on DeepSeek, outperforming HybriMoE and Fiddler by 42.1\% and 72.3\%, respectively. This advantage stems from PreSched's ability to minimize latency through cross-layer scheduling, leveraging prefetch benefits and coordinating compute/transfer more effectively than greedy strategies. 
At batch size 64, LayerScope attains 3.74 seconds on Qwen3-30B, reducing latency by 74.6\% and 46.7\% compared to Klotski and Fate. While Klotski’s multi-batch pipelining improves throughput, it activates more experts simultaneously, exacerbating I/O.
Fate lacks effective overhead masking and prefetch-compute parallelism, resulting in latency that scales linearly with batch size. For example, on Qwen3-30B, its latency increases by 4.6 times when the batch size increases from 4 to 64.

\textbf{Large-Expert Models:} Mixtral increases compute intensity,
yet I/O remains the bottleneck. LayerScope employs a dual-pipeline prefetch-compute approach that hides prefetch I/O behind computation. On V100 at batch size 64, it achieves 2.79 seconds, outperforming Fate and Klotski by 30.5\% and 53.9\%. Klotski’s latency plateaus at 5.82 seconds for batch sizes $\geq$ 32 as the activated expert count reaches its limit; LayerScope maintains flat latency via accurate prediction for early bandwidth release.

In summary, across architectures and batch sizes, LayerScope reduces latency by up to 68.2\%, 72.3\%, 74.6\%, and 42.1\% compared to Fate, Fiddler, Klotski, and HybriMoE, demonstrating its optimization effectively adapts to both “transfer-bound” and “compute-bound” scenarios.

\begin{figure}[t]
  \centering
  \begin{subfigure}[ht]{\linewidth}
    \includegraphics[width=\linewidth]{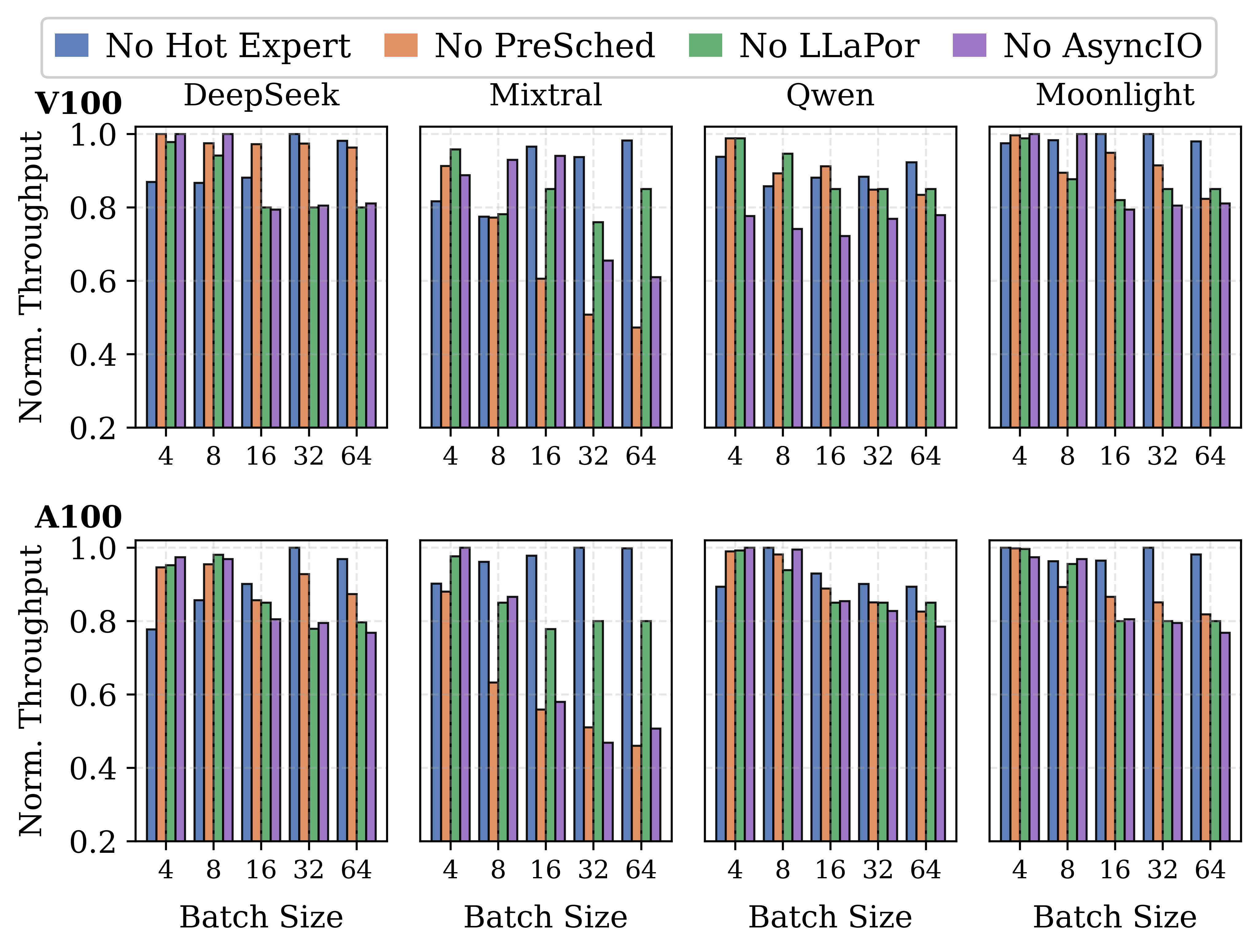}
  \end{subfigure}
  \caption{Throughput impact of ablating individual components in LayerScope.}
     \label{figure15} 
\end{figure}

\subsection{Ablation Study}

To quantify each module's contribution to end-to-end throughput in LayerScope, we perform ablation studies by individually disabling components and comparing against established baselines: 1) deactivating the Hot Expert Table removes expert preloading, testing extreme memory constraints; 2) deactivating PreSched reverts to a fixed policy of prefetching exactly two experts per layer, mimicking HybriMoE; 3) deactivating LLaPor enforces a greedy, layer-by-layer on-demand strategy; 4) deactivating AsyncIO restores PyTorch's default synchronous data transfer.

\begin{figure}[t]
  \centering
  \begin{subfigure}[ht]{\linewidth}
    \includegraphics[width=\linewidth]{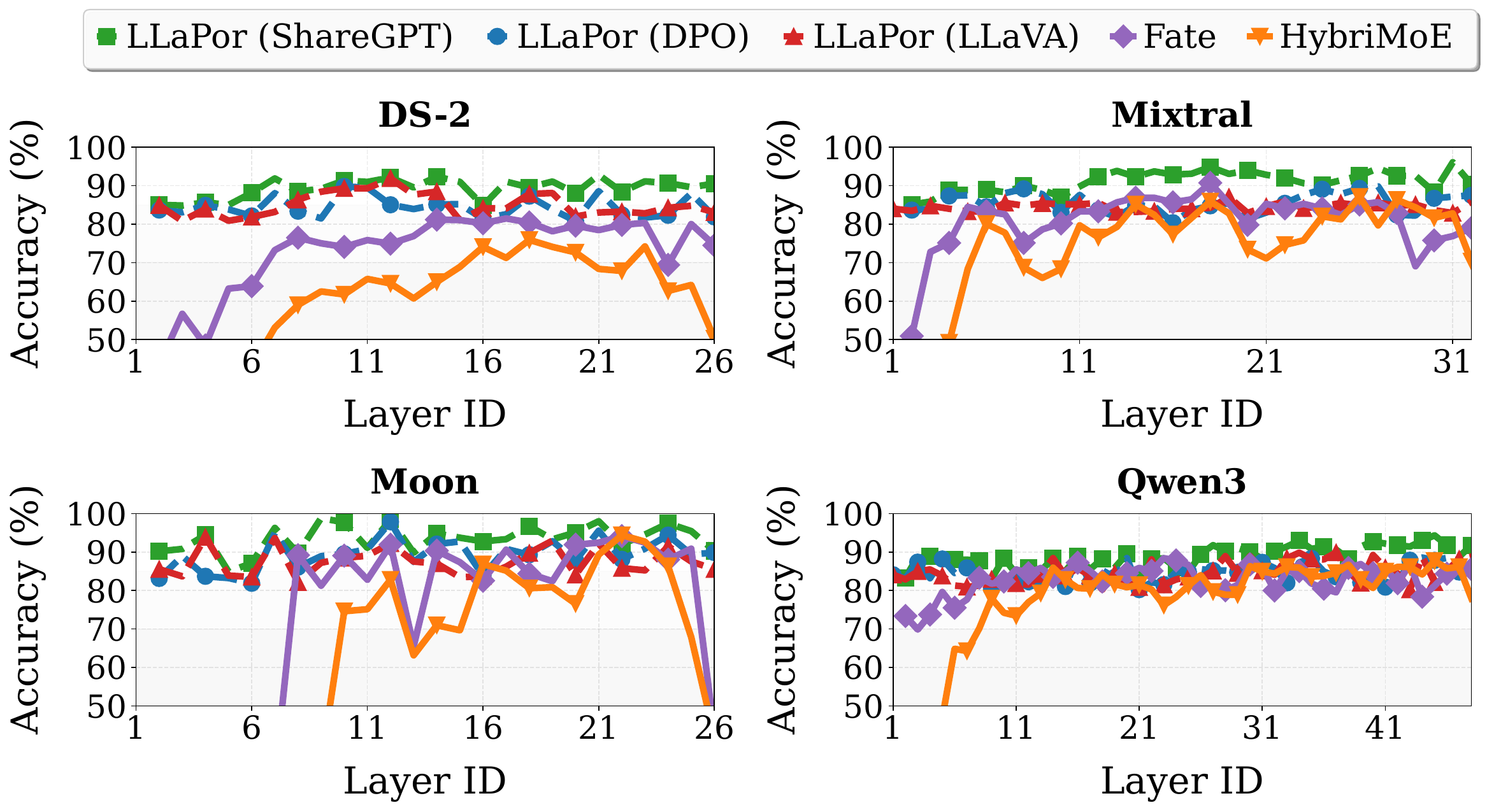}
  \end{subfigure}
  \caption{Prediction accuracy of LLaPor across different models and datasets.}
     \label{figure16} 
\end{figure}

The results in Figure~\ref{figure15} show that system bottlenecks shift with batch size. When the batch size is below 16, removing the Hot Expert Table reduces cache hit rates by 30–40\%. For the DeepSeek model on V100 at batch size 4, throughput decreases by 22.2\%, highlighting the critical role of the Hot Expert Table in reducing redundant transfers under light load.
As the batch size exceeds 8, computational costs on the CPU rise significantly, making prefetch timing crucial. Disabling PreSched on A100 leads to a 54.1\% throughput reduction for the Mixtral model at batch size 64, while removing LLaPor causes a 20.2\% decrease for the Moonlight model at batch size 32. Under medium to high loads (batch size $\geq$ 16), disabling LLaPor or PreSched on A100 results in more severe losses, as high PCIe bandwidth allows more prefetch operations to be scheduled during CPU computation, increasing the prefetch ratio. Disabling AsyncIO serializes expert transfers and computations, and Pytorch's  coarse-grained operations are over 70\% slower than AsyncIO’s fine-grained approach, leading to a sharp drop in PCIe utilization. This impact is particularly pronounced under medium to high loads; for example, throughput decreases by 49.3\% for Mixtral on A100 at batch size 64.

In summary, the ablation results demonstrate a clear dependency hierarchy: the Hot Expert Table ensures memory efficiency under light loads, but its contribution can be substituted by the optimizations of other components in large-batch-size scenarios, reflecting LayerScope's adaptability to extreme memory constraints. In contrast, PreSched and AsyncIO jointly optimize compute-I/O overlap at medium to high loads through scheduling and execution improvements, while LLaPor provides reliable future expert activation visibility. These latter three components are indispensable—removing any one of them results in 20–54\% performance degradation, validating the necessity and robustness of LayerScope's co-design.

\subsection{Predictor Accuracy}
\label{preacc}

To evaluate the generalization capability and prediction accuracy of LLaPor, we first sample 16\% of tokens from the ShareGPT dataset for offline training. Online testing is then performed on 32 randomly selected batches (each of 4096 tokens) from three public datasets: ShareGPT, DPO, and LLaVA. Prediction accuracy is assessed by comparing the experts selected by the gating network against those predicted by LLaPor and calculating the hit rate. 

\begin{table}[t]
\centering
\caption{Performance Evaluation of LLaPor.}
\label{tab:model_performance}
\resizebox{\linewidth}{!}{%
\begin{tabular}{l *{4}{>{\centering\arraybackslash}p{2cm}}}
\toprule
 & \textbf{DeepSeek} & \textbf{Mixtral} & \textbf{Qwen3} & \textbf{Moonlight} \\
\midrule
\textbf{Top2} & 100\% & 100\% & 100\% & 100\% \\
\textbf{Top3} & 97\% & 100\% & 99\% & 100\% \\
\textbf{Top4} & 94\% & 99\% & 97\% & 99\% \\
\textbf{Tokens} & 95\% & 97\% & 97\% & 98\% \\
\hline
\textbf{Memory} & 0.6-2.7 MB & 0.5-2.5 MB & 0.5-2.8 MB & 0.6-2.7 MB \\
\textbf{Time} & 0.48 ms & 0.15 ms & 0.12 ms & 0.48 ms \\
\bottomrule
\end{tabular}%
}
\label{tb3}
\end{table}

Results are shown in Figure \ref{figure16}. LLaPor’s prediction accuracy varies across models, aligning with the analysis in \S\ref{32}: it is positively correlated with inter-layer cosine similarity and expert selection correlation. Notably, LLaPor exhibits only an average accuracy drop of 4.3\% on DPO and 4.7\% on LLaVA versus its offline training dataset (ShareGPT). Crucially, for any model, even in low-similarity input/output groups, LLaPor’s accuracy never falls below 80\%, demonstrating strong generalization to domain shifts. Compared to Fate, LLaPor improves accuracy by 15\% to 68.4\%.

We further examine the impact of LLaPor on end-to-end operation. Table \ref{tb3} shows the "hot Top-4" prefetch accuracy: experts are ranked by token volume, and with a second-order sliding window (i.e., a prediction is considered correct if the Top-4 falls within the true Top-6), LLaPor achieves a Top-4 hit rate of no less than 94\%, corresponding to a token coverage error below 5\%, meeting PreSched’s precision requirement for CPU-side overhead estimation.
For resource usage, each model deploys one micro-predictor per layer, with per-predictor parameters as low as 0.5 MB. Middle-layer predictors have slightly deeper structures, with a maximum footprint of 2.8 MB per layer. During inference, single forward-pass latency remains stable between 0.12–0.48 ms, which can be fully hidden within gaps where the GPU awaits expert transfers, introducing no additional bubbles.

\subsection{CPU-GPU Parallelism}

\begin{figure}[t]
  \centering
  \begin{subfigure}[ht]{\linewidth}
    \includegraphics[width=\linewidth]{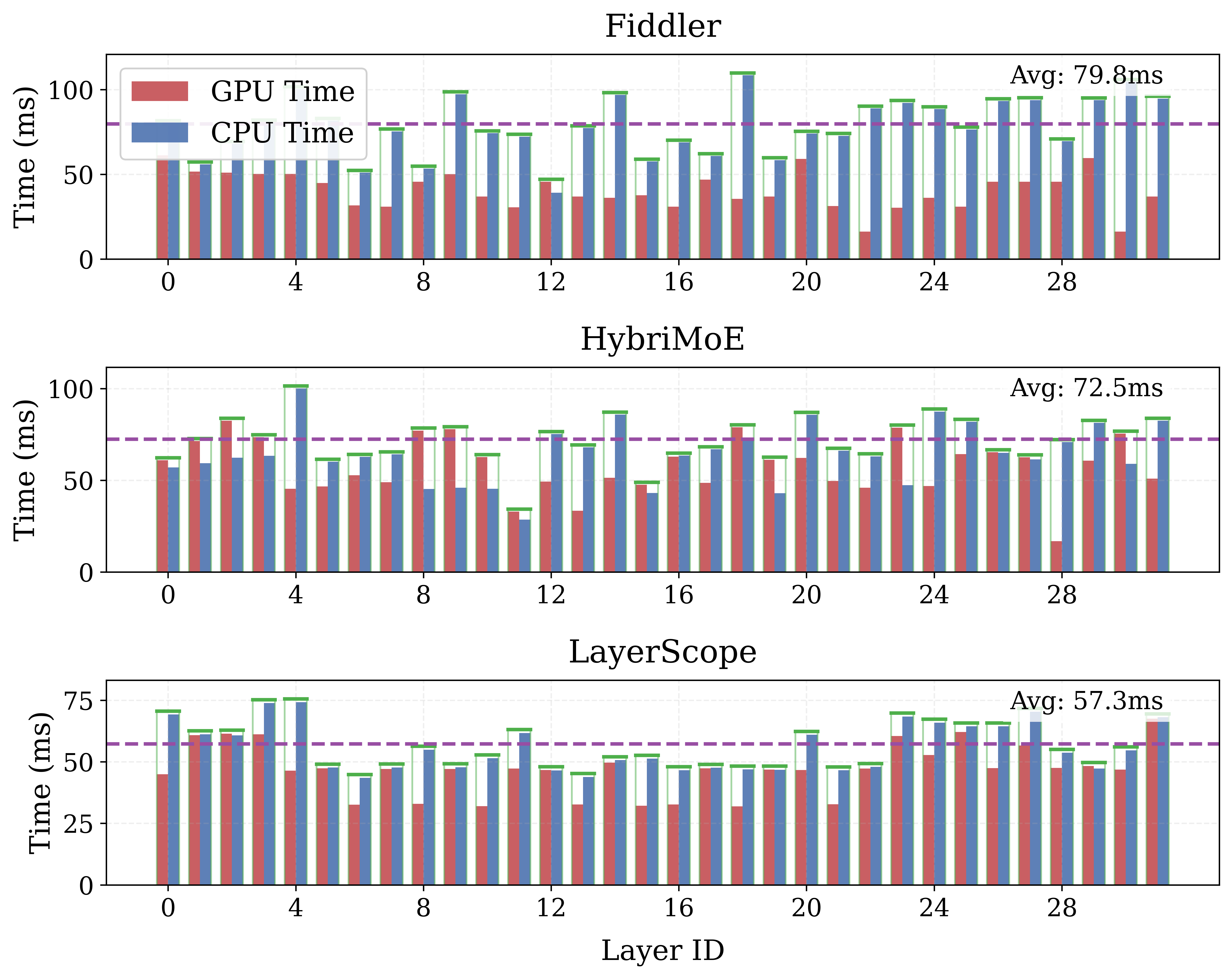}
  \end{subfigure}
  \caption{Per-layer time breakdown comparison between LayerScope and collaborative inference baseline methods.}
     \label{figure17} 
\end{figure}

The key to CPU-GPU collaborative inference lies in maximizing parallelism between the CPU and GPU. Figure \ref{figure17} shows the cumulative distribution of per-layer CPU and GPU thread overheads for Mixtral on A100. LayerScope confines the gap between the two within 8 ms for 50\% of the layers; the remaining layers exhibit a peak difference of about 14 ms, aligning exactly with the latency of one expert prefetch. This indicates that LLaPor and PreSched, through a single-layer lookahead mechanism, decouple prefetch overhead from the critical path, achieving precise overlap of CPU computation with GPU computation and transfer. Furthermore, LayerScope prevents high CPU thread overhead in layers containing numerous hot experts (mostly non-resident in GPU during initialization)—PreSched’s cross-layer scheduling anticipates such scenarios and reserves on-demand loading opportunities in previous layers for prefetching larger experts, thereby balancing the load. 
In contrast, HybriMoE does not account for prefetch effects when placing experts, often prefetching too many experts in the current layer and delaying on-demand loading for the next layer, resulting in higher GPU than CPU thread overhead (e.g., layers 9–10). Alternatively, it fails to leverage CPU-GPU overhead differences for sufficient prefetching or suffers from prefetch failures, leaving the next layer’s overhead unoptimized (e.g., layers 28–29). Fiddler, lacking the cold-hot expert queue, retains 70\% of hot experts on the CPU, causing average CPU thread overhead to be 1.9 times that of the GPU and yielding the lowest parallelism.
Experiments show that LayerScope, via prediction-driven cross-layer scheduling, compresses the CPU-GPU thread overhead difference to within a single prefetch granularity, significantly outperforming layer-by-layer scheduling strategies and providing a quantifiable upper limit for collaborative inference parallelism.





\section{Conclusion}

This work presents LayerScope to address the memory bottleneck and PCIe latency in MoE inference under resource-constrained environments. By jointly modeling the current layer and the prefetch-target layer, LayerScope reformulates expert on-demand loading and prefetching decisions into a global latency-aware expert placement optimization problem.
Three co-designed core components collectively address the associated challenges: LLaPor ensures accurate target-layer information through group-aware prediction; PreSched constructs a pipeline model incorporating I/O, GPU, and CPU execution within the cross-layer framework, and searches for latency-optimal scheduling strategies; AsyncIO provides system-level support to achieve full overlap between I/O operations and GPU computation while maximizing PCIe bandwidth utilization efficiency.
Evaluations on four MoE models show that LayerScope improves throughput by 141\% and reduces decoding latency by 74.6\% compared to state-of-the-art baselines, while maintaining stable scalability from batch size 4 to 64 on legacy servers.




\begin{acks}
This work is supported by the National Natural Science Foundation of China under Grant No.62421002 and No.U24B20151. Dezun Dong, Haojie Wang, Yongwei Wu, and Xiangke Liao hold joint appointments at Qiyuan Lab. Dezun Dong, Zhaoning Zhang and Haojie Wang are corresponding authors.
\end{acks}

\bibliographystyle{ACM-Reference-Format}
\bibliography{bib2}


\end{document}